\def\year{2022}\relax
\documentclass[letterpaper]{article} % DO NOT CHANGE THIS
\usepackage{aaai22}  % DO NOT CHANGE THIS
\usepackage{times}  % DO NOT CHANGE THIS
\usepackage{helvet}  % DO NOT CHANGE THIS
\usepackage{courier}  % DO NOT CHANGE THIS
\usepackage[hyphens]{url}  % DO NOT CHANGE THIS
\usepackage{graphicx} % DO NOT CHANGE THIS
\usepackage{natbib}  % DO NOT CHANGE THIS AND DO NOT ADD ANY OPTIONS TO IT
\usepackage{caption} % DO NOT CHANGE THIS AND DO NOT ADD ANY OPTIONS TO IT
\DeclareCaptionStyle{ruled}{labelfont=normalfont,labelsep=colon,strut=off} % DO NOT CHANGE THIS
\usepackage{multirow}
\usepackage{algorithm2e}
\RestyleAlgo{ruled} 
\SetKwComment{Comment}{\triangleright}{}

\usepackage{subcaption}

\title{Automated Play-Testing Through RL Based Human-Like Play-Styles Generation}
\author {
    Pierre Le Pelletier de Woillemont\textsuperscript{\rm 1,\rm 2},
    Rémi Labory\textsuperscript{\rm 2}, 
    Vincent Corruble\textsuperscript{\rm 1}
}
\affiliations {
    \textsuperscript{\rm 1} Sorbonne Université, CNRS, LIP6, F-75005\\
    \textsuperscript{\rm 2} Ubisoft La Forge, France\\
    pierre.le-pelletier-de-woillemont@ubisoft.com,
    remi.labory@ubisoft.com,
    vincent.corruble@lip6.fr
}

\begin{document}

\maketitle

\begin{abstract}
The increasing complexity of gameplay mechanisms in modern video games is leading to the emergence of a wider range of ways to play games.
The variety of possible play-styles needs to be anticipated by designers, through automated tests. Reinforcement Learning is a promising answer to the need of automating video game testing. To that effect one needs to train an agent to play the game, while ensuring this agent will generate the same play-styles as the players in order to give meaningful feedback to the designers.
We present \textit{CARMI}: a Configurable Agent with Relative Metrics as Input. An agent able to emulate the players play-styles, even on previously unseen levels.
Unlike current methods it does not rely on having full trajectories, but only \textit{summary data}. Moreover it only requires little human data, thus compatible with the constraints of modern video game production.
This novel agent could be used to investigate behaviors and balancing during the production of a video game with a realistic amount of training time.
\end{abstract}

\section{Introduction}
\noindent In the video game realm, the main goal of Reinforcement Learning (RL) has usually been to achieve superhuman performances \cite{AlphaZero} \cite{AlphaStar}.
%on board games \cite{AlphaZero}, simple video games \cite{mnih2013playing} or even highly complex ones \cite{AlphaStar}.
%as evidenced by the work on board games with AlphaGo Zero \cite{AlphaZero}, on simple video games such as Atari with DQN \cite{mnih2013playing} and on complex video games with AlphaStar \cite{AlphaStar} on StarCraft II.
%While providing a challenging experimental common ground for researchers, this focus of superhuman performances can have unforeseen consequences when applied to the video game industry. One of which is that more and more players feel discouraged by the fact that an algorithm can beat the best players in the world. For example Lee Sedol, the 18 times Go champion, announced his retirement " With the debut of AI in Go games, I’ve realized that I’m not at the top even if I become the number one through frantic efforts. I kind of felt powerless".
Another interesting application for RL is the pursuit of human behavior \cite{Strong_human_like_kl_regu_search}, in order to perform automated game testing. %, for difficulty assessment for example. 
These automated tests can have varying objectives: find the most resource consuming assets of the game or flag possible  unintended exploits in the gameplay. In our case we wish to perform automated tests in order to help assess more accurately the difficulty of the game.

The results of these automated tests are most useful to the designers during the production phase of the game, during which there is very little human data available (only the occasional human play-tests). Moreover, this data is rarely in the form of full trajectories but rather in the form of \textit{summary data}, i.e. in the form of a few key metrics, or \textit{play-modes}, %\cite{PersonasInGame2008}
 that are of most interest to the designers (e.g. number of shots fired during a game).
Indeed, tracking and storing full players' paths (state-action tuples) is a big constraint to put on developers, especially in the early stages of the production, as it involves issues of bandwidth and computational resources. 
%It can prove unreliable due to crashes.
Whereas summary data is cheaper to acquire and to store.
%Moreover, the changes between two versions of the game can lead to the necessity of having to disregard the trajectories of past play-tests, while the summary data of past trajectories might still be useful.
Furthermore, the pipeline for tracking and storing summary data is usually the same during the production of the game, as it is post launch.
In short, logging summary data is cheaper, easier to implement and is most likely already being done by the developers.
In this work, the available human data is assumed to be very limited and in a summarized format: no full trajectories, only a few key metrics, so as to fit with the constraints of game production.
%This means that we do not have the full trajectories (state-action tuples) of the players' games, but have instead some key metrics measuring their activity within the game for each level they played (e.g. the number of shots fired during a level).
%This is necessary to make this approach usable during the production phase.
%And because of this assumption, only a small human dataset (corresponding to occasional play-tests) in the form of \textit{Summary Data} (SD) is available. 

To assess the difficulty of a game, distinguishing between performance (i.e. success) and play-style is important: a game can be perceived hard for a subset of the players because of their style of play, not because of the game itself. So the performances relative to the play-style should be reported to the designers, as well as the overall difficulty, so as to help them make informed decisions.
%The play-style is the manner in which a person plays.
A play-style is measured using key metrics which capture different aspects of the gameplay. %: for example the frequency of shots fired during a game.

We present here a Configurable Agent with Relative Metrics as Input (\textit{CARMI}) agent, which aims at generating a continuum of play-styles, fitting with the players distribution, making the sampling of human-like play-styles straightforward, even on levels previously unseen by the agent and the players.
This agent is configurable with respect to the metrics used to define the play-styles and is obtained through a single RL training loop.
%We develop a method for training an agent capable of generating human-like play-styles
%Additionally, we propose an inference strategy to generate human-like play-styles on newly unseen levels, using only summary data gathered during a play test with few players.
%
%The contributions of this work are two-fold. 

%This approach rests on the definition of a play-style.
%Like in \cite{StarcraftClustering}, 
The definition and measure of a play-style is done on summary data (e.g. the number of shots fired and the number of stabs made with a knife during a game). However, here it is normalized level by level, using the players distribution. Therefore the play-styles are defined not as absolute but as relative to the players population, making them independent from the levels and their designs.
Take for example a play-style characterized by "a high use of a riffle over a knife". Instead of measuring "high use" as "X times during a game" it would be measured "X times more than the other players", making the definition of this play-style applicable no matter the level.
%Take the example of the play-style characterized by a high use of the rifle. Instead of measuring this play-style as "use the rifle X times during the game", this play-style would be measured as "use the rifle X times more than the other players", making the definition of this play-style applicable no matter the level.
%For example, instead of defining a play-style as "the players who use the rifle X times during the game", this play-style would be defined as "the players who use the rifle X times more than the other players", making the definition of this play-style applicable no matter the level.
%Indeed, on one level using the rifle X times may be considered a lot, but it might be considered little on another level. Whereas using the rifle X times more than the other players will always be defined and measured the same way.
%This means that this approach could be used even if the generation of a level is a stochastic process, either because the game is non linear, or even because the game uses some form of procedural generation.

The agent learns a policy conditioned to matching a desired play-style. This is done by giving the desired relative summary data $z$ as input to the agent, and building a reward function that orients the agent towards matching this $z$.
It is the main contribution of this work: training an agent conditioned by a play-style which is defined in relationship to human play-styles distribution. This means that the agent learns the distribution of the players in the play-style space solely through the reward function.
%This allows a single training procedure resulting in a single configurable agent, w.r.t. $z$. From the agent point of view, the definition and the measure of the wanted play-style are one and the same: $z$.
%In the case of stochastic environments (which most games are) the agent is trained to take this stochasticity into account when generating a play-style.
Proceeding this way produces two results. 
First, it ensures that the agent is able to cover at least the same space as the human players in the space of the play-styles.
Second, the independence between the play-styles and the level design allows the agent to easily generalize the play-styles to newly unseen levels. %It is the main contribution of this work.
This is due to the fact that the agent has learnt to associate any value of $z$ to a behavior, no matter the level, which allows better generalization. 
The new direction presented by this paper results from the capacity to generalize human summary data to new levels using few human data, to better fit with the constraint of the industry.
%This is due to the fact that the agent has learned that the same value of $z$ might results in very different absolute metrics depending on the level design, because $z$ is relative.
%shooting as much as the  players who shoot the most might mean shooting twice in one level 

We can therefore use this agent to play-test new levels, using play-styles defined from human data collected on previous levels. And because the agent is conditioned using a very interpretable input $z$, it could also be used for automated play-tests using designers' hand-made play-styles, since each dimension of $z$ corresponds to a specific metric.

In this paper we first give a more formal definition of play-styles as well as the existing methods that aim at learning various personas and those aimed at modeling human behavior.
We then introduce the environment: a turn based strategic shooter. And finally compare our results with the existing method CARI \cite{CARI_paper}.

\section{Background}
\subsection{Play-Styles As A Combination Of Play-Modes}
%Canossa and Drachen 
\citeauthor{Canossa2009PatternsOP} adapt the "persona" framework introduced by
%Alan Cooper 
\cite{CooperPersonas} in the field of Human Computer Interaction.
%They defined personas within a video game as the expression of the persona but within the limited space of a specific video game and called it \textit{play personas}. 
%Tychsen and Canossa 
\citeauthor{PersonasInGame2008} make the distinction between play-mode, play-style and play-persona. It is a distinction based on the level of data aggregation. 
A play-mode is one or a few discrete metrics, within the same overall group or type of metrics.
%Play-mode "refers to the behavior of a player with respect to one or a few discrete metrics, within the same overall group or type of metrics". 
From there, play-style is defined as "a set of composite play-modes".
And finally play-personas represent the "larger-order patterns that can be defined when a player uses one or more play-styles consistently". The focus of this paper is on the simulation of human-like play-styles through the generation of behaviors yielding play-modes similar to players'.
%To that effect absolute play-modes are not used, instead the play-modes are normalized based on the distribution of the players on these play-modes. Therefore play-styles are not defined as absolute but rather as relative to the other play-styles observed amongst players.

%Not all play-styles perform at the same level.
%In the case of a game focused around infiltration and stealth for example, a player who likes to rush in and charge head-on will most likely not perform well compared to a stealth player.
%Those observed play styles result from the analysis of quantitative players' data gathered via telemetrics \cite{PersonasInGame2008}, for instance via clustering algorithms, and indicate the ways the game is played.
%they might coincide or not with the play-styles intended by designers, and can prove to be efficient (in terms of performance level) or not.
%Most games today, especially the ones coming from major studios, have built-in tracking mechanisms allowing some form of clustering on the players to be done, based on their behavior within the game. It is usually done on key metrics, or play-modes %\cite{StarcraftClustering}
%\cite{ClusteringWildlands}.
%This combination of tracking and clustering is done today on many of the big budget games \cite{ClusteringWildlands}. It allows designers to know the main ways players play their game.

\subsection{Automated Play-Testing}
Recent work as been done to develop automated play-tests using machine learning (ML) approaches, more specifically using agents trained through (Deep) RL.
%It is important to differentiate these works based on their objectives.
There are two main categories of test: technical tests (e.g. frame rate, bugs) and gameplay tests (e.g. difficulty assessment, game consumption analysis). Our approach focuses on difficulty assessment for the experiments while taking into account the diversity of approaches and resulting play-styles.

RL can be use to improve game testing \cite{Automated_game_testing_using_DRL} in order to find unintended exploits in the video game.
It can also be used to measure what is achievable in the game. For example, \cite{CCPT} train an agent to uncover what is possible in the environment, but should not be, in order to flag bugs and glitches for the designers to fix.
\citeauthor{smartnav} trained an agent to navigate complex 3D environments, which can be used for Non Playable Character (NPC) development with complex navigational skills, or in order to measure the feasibility of procedurally generated goals. 
%RL can also be used to perform game difficulty balancing: \cite{andrade_hal_01493239} and \cite{andrade_hal_01492622} used RL trained agents to dynamically adjust the difficulty of the player's opponent. 

However, in this work we are interested in training an agent that can inform us about the difficulty of the game. Therefore, this agent must produce diverse human-like play-style in order to provide meaningful feedback.
%However, in this work we are interested in training an agent capable to reproduce human-like play-styles on newly unseen level in order to help the game designers assess the difficulty of their game more accurately.
%Although the use of player-facing RL agents is a very exciting and cutting edge feature to be developed, our work focuses on the use of RL agent to produce automated tests to help the production of a video game, not to be shipped with the game at release.
%In order to provide the game designer with such an agent for game testing purposes, one must incorporate the different play-styles in this model.
%The idea of using trained agent to automate play-testing has been explored in depth by Holmg{\aa}rd et al. through archetypal personas generation \cite{holmgard2014Generative}, \cite{holmgard2014Evolving}, \cite{holmgard2018automated}. In their work they mostly focus on archetypal agents and their alignment with players' decisions. In addition they also showcase how those stereotypical play personas can be used to improve content generation. An extension of their work has been proposed in \cite{CARI_paper} to non-archetypal personas by removing the necessity of one model per play-style by fitting one model on a continuum of play-styles. 
%Our work builds on this approach, keeping the one agent framework, but training it in a a way that will make it closer to human players and making the inference strategy on new levels simpler.

\subsection{Diverse Human-Like Behavior} \label{sec:Human_Like_Behavior}
To approximate human behavior, Inverse Reinforcement Learning (IRL) \cite{ng2000algorithms} can potentially be used.
The 2 main drawbacks of IRL methods are the homogeneity assumption in the trajectories and the quantity of the data necessary for such approaches to yield convincing results.
%Indeed most methods rely on having a large number of trajectories to sample from.
In our case, neither full trajectories nor large dataset are available.
%We assume not only to have very little data but we also assume that these data are summarized.
\citeauthor{IRL_SD} developed an approach to perform IRL using solely summary data to alleviate this problem. However they assume homogeneity in the data, in the sense that all the data comes from the same expert. 
This homogeneity assumption goes against the idea of generating varying, and therefore heterogeneous, play-styles.
%Moreover, assuming that all the data came from experts doesn't allow the use of this method for balancing purposes.

% \noindent \citeauthor{christiano2017deep} \cite{christiano2017deep} use human feedback to drive the training of the agent. While having a human in the training loop, will be a strong requirement to achieve the human-like model we seek, it does not address the play-style dimension of this human-like model. In this paper no human feedback is being used.

%There have been many attempts to tackle this idea of play-styles.
There are a few possibilities to insert heterogeneous play-styles into an agent.
To that effect, \citeauthor{holmgard2018automated} have demonstrated that shaping the reward (\citeyear{holmgard2014Generative}), fitness (\citeyear{holmgard2014Evolving}) or utility function (\citeyear{holmgard2018automated}) produces variety in the style of play of the agents. 
%To that effect \citeauthor{holmgard2018automated} use RL (\citeyear{holmgard2014Generative}) , evolutionary (\citeyear{holmgard2014Evolving}) and MCTS (\citeyear{holmgard2018automated}) agents. They demonstrated that reward, fitness or utility function shaping achieves variety in the style of play of the agents. 
Moreover they showcased that archetypes (i.e. stereotypical play-styles) can be a good low-cost, low-fidelity approach to automated play-test. %\cite{holmgard2014Evolving}.
%To better fit with the language used in the video game industry the term \textit{archetypes} is used instead of stereotypical play-styles.
Because each archetype requires its own training and its own reward function, they limit themselves to only 4 archetypes.
%The main issue with their approach is the limitation to only 4 archetypes. This is due to the fact that each archetype requires its own training and its own reward function. 
%This reward function is usually designed by a data science practitioner, and therefore might not represent anything close to human play-styles.

\citeauthor{CARI_paper} (\citeyear{CARI_paper}) solved this issue by training a \textit{CARI} agent where the coefficients of the reward function are given to the RL agent as input. Concretely, they train a policy $\pi(a|s,w)$ where $w$ are the coefficients of the reward function: $r = w \cdot  \theta$, where $\theta$ represents all the events which induce a reward signal. 
Additionally one must define $W$, the intervals within which the $w$ will be sampled during the training.
Thus creating a continuum of available play-styles to sample from at inference time (simply using varied $w$). This approach allows to train one single model rather than multiple ones. It moves the problem of selecting the proper $w$ to generate the desired play-style after the training rather than before, it does not however remove it.

The CARI approach suffers from two main issues. The first is the absence of guarantee that the space of generated playstyles covers well enough the set of human play-styles. The second is that even if the CARI generated play-styles do include the players' play-styles, there is no straightforward way to know which reward coefficients $w$ to select in order to simulate the human-like play-styles. This is due to the fact that the process still relies on finding a good combination of reward coefficients that will hopefully generate the desired play-modes.
Both these issues are solved here by directly giving the target values of the desired play-modes as the goal to the agent as part of its input state: replacing $w$ by a $z$ which encodes a desired play-style. These objectives are encoded and sampled using the distribution of the play-mode over the players population.
%In addition, the CARI approach \cite{CARI_paper} does not ensure that the model can simulate the same behaviors as the players'.
%While allowing the generation of a wide variety of play-styles, the CARI approach \cite{CARI_paper} does not ensure that the space of generated playstyles covers well enough the set of human play-styles. Moreover, even if the CARI generated play-styles do include the players play-styles, there is no straightforward way to know which reward coefficients to select in order to simulate the human-like play-styles.
%This is due to the fact that the process still relies on finding a good combination of reward coefficients that will hopefully generate the desired play-modes.
%This issue is solved here by directly giving the target values of the desired play-modes as the objectives to the agent as part of its input state, instead of the reward coefficients. These objectives are encoded and sampled using the distribution of the play-mode over the players population.
This agent still allows access to a wide variety of play-styles but also ensure that the generated play-styles do in fact encompass the humans' ones.
%Moreover, it becomes straightforward to sample from these play-styles to find the ones which represent best the players distribution.

%We do not however claim here that this agent will behave like players on the full trajectories, only that it will generate the same key metrics values as players, thus generating the same play-styles as they are measured: by a set of metrics.

However, there is no guarantee that the players' and the agent's full trajectories look the same, for that we would need full human trajectories to train on, which we assume to not have access to during the the production of the game. We are not claiming to generate human-like behaviors, but human-like play-styles, as they are measured : using summary data. The goal is to produce summary data that are player-like enough to take well informed decisions, on new levels.

\subsection{Goal-Conditioned Reinforcement Learning}
Goal-conditioned RL has been used in video game testing in different ways. For example \cite{constraint_RL}, frame their problem as constrained RL, to create an agent with the desired behavior. Their approach relies on a main goal to achieve and several constraints to fulfill, and are automatically weighed throughout training. This approach does not allow the emergence of heterogeneous play-styles, necessary in our case, since the play-style is the goal to achieve, not a constraint to be met.

Goal-conditioned RL has been extensively studied \cite{goal_cond_rl_old} \cite{UVFA_goal_cond_value_function} \cite{Goal_RL_subgoals}. The framework is usually in the form of a policy which given a state $s$ and a goal $g$, predicts actions which lead to the goal. The goal is usually a certain state of interest in the environment (e.g. a destination). In our case the goal is the encoded play-style $z$, relative to the players distribution. It is not a state in the environment, but a behavior to adopt.
Moreover, this behavior is not absolute, but relative to the players, and should be matched on different levels. The only information about the distribution of the players given to then agent is thought the reward feedback.

\section{Proposition}

Our goal is to learn a play-style conditioned policy which generates trajectories yielding summary data as close as possible to a given objective. This policy is then used to emulate human-like play-styles, on previously unseen levels.

\subsection{Notations}

In our environment, let $\tau = (s_0, a_1, s_1, ..., a_T, s_T)$ denote the trajectory of a full episode. % starting at $t=0$ and ending at $t=T$.
%Let $\tau_t$ denote a trajectory until time-step $t$.
We introduce the summarizing function $ \psi $ which takes as input a trajectory and returns a set of $M$ summary data (e.g. number of shots and stabs made): $\psi(\tau) \in \mathbf{R}^M$.
In this work, we assume to have access to the summary data of $N_P$ players over $N_L$ levels of the game: $\{ \psi(\tau_{l,p})  \}_{l \in [1:N_L], p \in [1:N_P]}$. Additionally, $\mu_l$ and $\sigma_l$ represent the mean and standard deviation of the players summary data on level $l$, such that $\psi(\tau_{l,.}) \sim \mathcal{N}(\mu_{l}, \sigma_{l})$.
We also note $\psi_l(\tau) := \frac{\psi(\tau) -\mu_l}{\sigma_l}$, the summary data on level $l$, normalized by the players distribution on that level (assuming a Gaussian distribution).

We introduce the policy $\pi_z (a|s) := \pi(a|s,z) $, with $a$ the action, $s$ the state and $z$ some additional information (e.g. a play-style to emulate in our case or the reward coefficients in CARI) given to the agent, with $z \sim \mathcal{P}_z $, here we assume that $\mathcal{P}_z =  \mathcal{N}(0,1) $. In our case we wish for $z$ to represent $\psi_l(\tau)$.
Our goal is for $\pi_z$ to generate $\tau_{\pi_z}$ yielding $\psi_l(\tau_{\pi_z})$ as close as possible to $z$.
The reward function reflects this objective by being proportional to $-d_t := -\Delta(\psi_l(\tau_{\pi_z, t}); z)$, the current distance between the goal $z$ and the agent. Where $\tau_{\pi_z, t}$ is the trajectory generated by $\pi_z$ until time-step $t$. 
By formulating the problem this way, we can then use any RL algorithm to solve this MDP and generate an agent with a policy $\pi \in argmin_\pi 
\{\mathbf{E}_l [\mathbf{E}_z [ \mathbf{E}_{\tau \sim \pi_z} [ \psi_l(\tau)-z ]]] \}$.

\subsection{Relative Metrics As Input}
The way the metrics are encoded, when given to the agent as goals, is key here.
%The first contribution of this work is the way the metrics are encoded when given to the agent as objectives.
Instead of aligning $z$ (the agent's goal) to the absolute metrics values $\psi(\tau)$, it is aligned to the relative ones, w.r.t. the players distribution on that particular level: $\psi_l(\tau)$.
%Instead of giving to the agent the absolute metrics $\psi(\tau)$, the model perceives them as relative to the players distribution on that particular level $\psi_l(\tau) = \frac{\psi(\xi)-\mu_l}{\sigma_l}$.
It moves the issue of playing with a given play-style from a hard to define absolute point of view, into an easier relative one.
It is difficult to define what can be considered a \textit{high} frequency of shots, it is however easier to say if a given frequency of shots is among the \textit{highest} over the players population.

Additionally, it improves greatly the usefulness of the agent at inference time, especially on new previously unseen levels.
Giving the absolute metrics as the objective (i.e. $z \approx \psi(\tau)$) would have the agent learn how to achieve that specific goal, no matter the level design. During inference the agent would then still generate these exact metrics values, if the level allows it. This would render the agent quite useless for any automated testing procedure: on a new level the agent is tasked with shooting twice and it does it. We did not learn much, except that shooting twice on this new level is possible.

Instead giving the normalized metrics, relative to the players, as the objective (i.e. $z \approx \psi_l(\tau)$) means that the agent needs to learn to adapt its behavior to fit with the portion of the players represented by $z$.
Since we assumed the data to be normally distributed and given than $P(X \sim \mathcal{N}(0,1) < -1.96) = 0.025$, at inference time giving the target value $z = -1.96$ to the agent means shooting as much as the $2.5\%$ of players that shoot the least. 
After playing one level with this $z$, the  absolute number of shots done by the agent is then reported to the designers, which would correspond to the number of shots done by this portion of the players, had they played that new level.
%This would be a great tool for automated testing. The agent is not learning how to achieve certain metrics values but instead how to generate the same metrics values as the whole players population.

%Moreover, the key play-styles defined later through clustering, in order to devise a sampling strategy, when fitted on the normalized metrics are then independent of the level design. This means that this clustering can be done on the players play-styles on train levels and it will hold on test levels. This is due to the fact that the clustering uses values that are independent to the level design.

\subsection{Configurable Agent}
%Unless a clustering is done on the players, there usually is as much possible play-styles as there is players. Training one model per play-style is not feasible, even doing one model per cluster's centroïd is not.
The more complex the game the more numerous the play-styles, thus the more numerous the number of models to train. In order to solve this issue a single model is trained, much like in \cite{CARI_paper}.
%Doing so would also allow the agent to emulate new un-anticipated play-styles, even after having already trained the agent.
The objective $z$, which correspond to the normalized metrics $\psi_l(\tau)$ are given inputs to the model. 
%Algorithm \ref{alg:CARMI_one_ep} details how this is done.
Proceeding this way makes the agent configurable: the play-style we wish the agent to adopt can be chosen after training, at inference time. We only need to train a single model for all play-styles.
At the beginning of each episode a new $z$ is drawn and given as input to the agent. The agent has then the objective of matching this value by the end of the episode.

%\begin{algorithm}
%\caption{Run an episode with CARMI: $RunEpisode(\pi, z) = \tau_{\pi_z}$ }\label{alg:CARMI_one_ep}
%\KwData{$\pi$ CARMI policy; $z$ normalized summary target}
%\KwResult{$\tau_{\pi_z}$ full trajectory obtained by CARMI under $\pi_z$}
%$\tau_{\pi_z, 0} \gets \{ (s_0)\}$ \;
%$d_0 \gets \Delta( \psi_l(\tau_{\pi_z, 0}), z)$ \Comment*[r]{ {\small Initial distance to objective}}
%$t \gets 0$ \;
%\While{Episode is not finished} {
%    $s_t \gets env_t(state)$ \Comment*[r]{{\small Get state}}
%    $a_t \gets \pi(s_t, z)$ \Comment*[r]{{\small Compute action}} 
%    $s_{t+1} \gets env_t(a_t)$ \Comment*[r]{{\small Execute action}} 
%    $\tau_{\pi_z, t+1} \gets \tau_{\pi_z, t} \cup \{(a_t,s_{t+1})\} $ \Comment*[r]{{\small Update trajectory}} 
%    $d_{t+1} \gets \Delta(\psi_l(\tau_{\pi_z, t+1}), z)$ \Comment*[r]{{\small Distance to objective}} 
%    $r_{t+1} \gets d_t - d_{t+1}$ \Comment*[r]{{\small Reward}} 
%    $t \gets t+1$ \;
%    }
%$T \gets t$ \;
%$r_T \gets -d_T$ \Comment*[r]{{\small Reward: sparse feedback}}
%\end{algorithm}

The reward function used to this end is defined as
$r_t = d_{t-1}-d_t + [-d_t ]_{t=T}$.
%$r_t = d_{t-1}-d_t$.
%Where $d_t = \Delta(\psi_l(\tau_{\pi_z, t}), z)$ with $\psi_l(\tau_{\pi_z, t})$ the current normalized summary data of the agent and $\Delta$ representing any distance. In our case $\Delta$ is the $L_1$ distance.
%A few things are worth noting about this reward function.
If the agent does an action bringing it closer to its objective it will receive a positive reward corresponding to how much closer it got, much like in a navigation problem.
The same applies if the chosen action moves it away from the objective. The objective of the agent is to end the episode as close as possible to $z$. 
This is why at the end of the episode the agent perceives a negative signal equal to how much distance to the target is left. Moreover, $\sum_{t=1}^T r_t = (d_0 - d_T) - d_T$: the cumulative reward on the whole episode is equal to the distance "\textit{travelled}" towards the objective minus the distance left at the end.

\section{Experimental Setup}
\subsection{Game Environment}

%Peut etre + aller en detail sur les type d'ennemis, le fait qu'il y a des hauteurs, des teleports
%Egalement présetner les données joueurs

The game environment used in this work is the same as in \cite{CARI_paper}. This environment presents a few advantages. It is complex enough to be able to have different play-styles and some of the challenges that come with training in a complex video game, but simple enough to have moderate computation requirements. Moreover it is developed using Unity's ML-Agents \cite{Unity_Ml_agent}, which allows to control the agent either with the Python programming language (for the RL aspect) or with a controller (for the human player aspect).
%This allows to easily organize play sessions during which players will play in place of the agent. These play-sessions provide the set of players metrics: $\{ \psi(\tau_{l,p})  \}_{l \in [1:N_L], p \in [1:N_P]}$.

%There exist already video games that have been used to train RL algorithms, but none that could really be of use for our particular problem of varying play-styles generation. The most famous one is the Atari suite \cite{Bellemare_2013} which, while being very useful due to the vast variety of games, is limited notably by its simplicity: it is not the type of video game where would emerge significantly varied and numerous play-styles. For this to emerge, one needs a more complex game. A good example is StarCraft II \cite{AlphaStar} where the vast complexity of the game offers many different approaches. Unfortunately the main drawback of this game is that it is actually too complex, requiring a vast amount of computation, and very long training time. We needed something complex enough to be able to have different play-styles and some of the challenges that come with training in a complex modern video game, but simple enough to have moderate computation requirements. 
\begin{figure}[]
\centerline{\includegraphics[width=\linewidth]{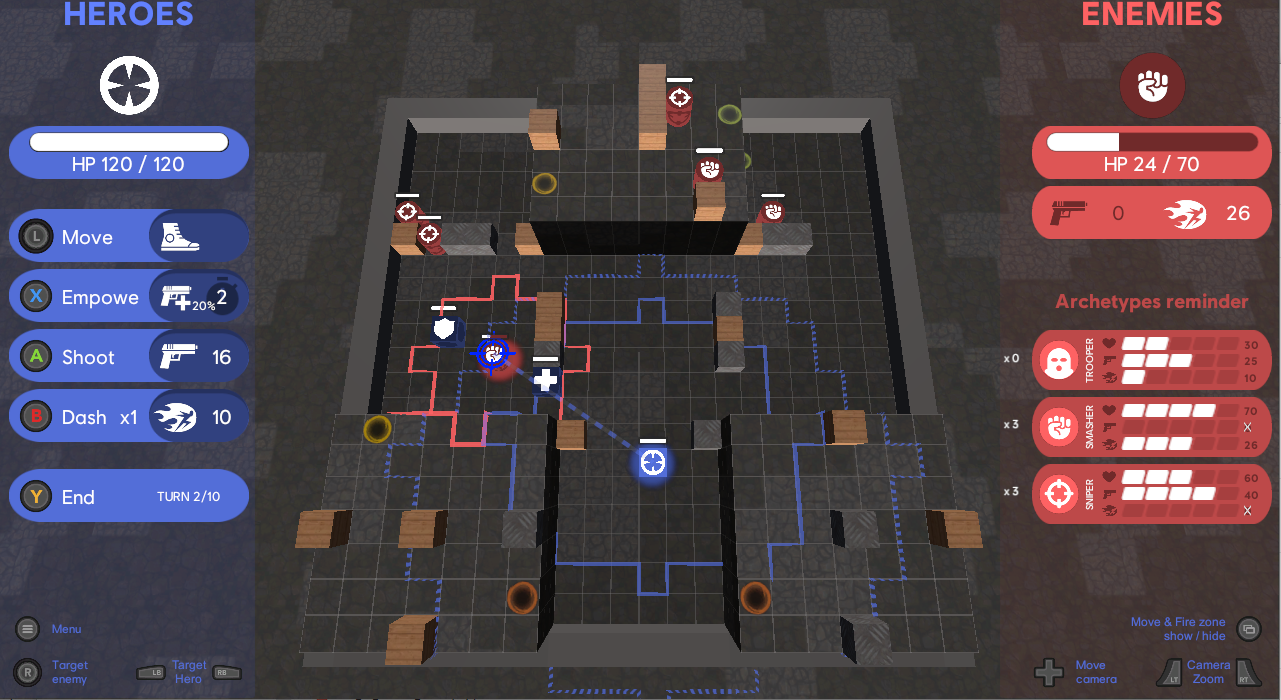}}
\caption{Screenshot of the Video Game}
\label{fig:Capture_proto}
\end{figure}

%\noindent This is the reason why this video game environment is used: a discrete, turn based, shooter-strategic, see Fig. \ref{fig:Capture_proto}.
This video game environment is a discrete, turn-based, shooter-strategic, see Fig. \ref{fig:Capture_proto}.
In this video game, two teams fight to the death on a 3D cell based board.
%of size 20x20x3. Each level consists of a base topology: whether flat or with 3 different heights levels. 
This game is inspired in its gameplay elements from the "Mario + Rabbids" video game, developed by Ubisoft.
The movements can be done in any directions toward an empty cell.
Some portals (the circles in Fig. \ref{fig:Capture_proto}) are spawned across the board to allow characters to take short-cuts.
%Every cell on the board is reachable through the use of these portals. 
In addition the board also consist of a series of covers behind which the characters can hide to avoid getting shot at.
%These covers are laid randomly on the board.

%One team, composed of 3 heroes, is controlled by the agent trying to learn or by the player, and the other team is controlled by hand-crafted decision functions, designed by us like any Non Player Character (NPC) would be in most video games. The agents controls the 3 heroes simultaneously: it is not a case of multi-agents, but a case of a single agent controlling multiple characters.
%Each team executes actions during their respective turn, they can move, deal damage from close (knife) or long (rifle) range, or apply a shield. Each character (hero or enemy) can move as much as needed during its turn, but cannot move anymore once it has used its rifle. The turn ends when no more action are feasible (on the enemies side) or whenever the agent decides to skip it (on the heroes side).
A team of 3 heroes (in blue), controlled by an agent (or a human, during a play-test), face a team of a varying number of enemies (up to 7, in red) controlled by hand-crafted behavior trees, designed by us like any NPC would be in most video games, for a maximum of 10 turns.
%The shield will absorb one attack before disappearing, and then it won’t be available for 2 more turns.
Every hero character has the capacity to move, shoot and stab (i.e. melee attack). They are all defined by a set of basic statistics: their health, their range of movement and of fire and their damages. Additionally, each hero a \textit{super capacity}. One has the power to heal nearby allies, another has the ability to apply an empowerment to nearby allies thereby increasing the damage caused by their attacks for one turn and the third hero can apply a shield that will block one attack. Each of these \textit{super capacity} has a two turns cool-down.
%There are three types of enemies: a sniper, a trooper and a smasher. The sniper cannot stab but has a high range of fire. The trooper can stab and shoot but does not have a lot of health. The Smasher cannot shoot but has a lot of health.

It is a turn-based game, meaning that when one of the team is playing, the other one is frozen in place. A full turn is not a single time-step, rather a full turn is many time-steps.
%They can move around as much as needed inside their area of movement, for example if the hero has a range of movement of 4 squares, then during the heroes’ turn, this hero can move anywhere at a maximum distance of 4 square from where it started on that turn.
%The same principle applies to its range of fire, each hero can shoot on any enemy within its range of fire. Once a hero has fired, it can no longer move during that turn. Each hero has one shot and one stab available during each turn.
So for example during one turn, the agent (controlling the whole hero team) can move around with hero number one, stab and shoot an enemy with the same hero, then shoot another enemy with hero number 3, and use the shield of hero number 2 before skipping the rest of its turn, effectively starting the enemy's turn. Each character has a maximum of one shot and one stab per turn. Once a character has shot it cannot move for the rest of the turn.
The game ends once one of the two teams has been fully destroyed or if the game reaches the 10 turns limit, in which case the game is considered to be a draw.
%One important property is that you can move as much as you wish any given hero (within its range of movement), but once this hero has fired, it cannot move anymore.

The state returned by the environment is twofold: an image-like segmentation map of size 20x20x3 indicating what object is inside each cell (hero, enemy, cover, portal or nothing at all) and an array comprising the rest of the information needed: the number of turns left, the current stats for each hero and each enemy
%(health, range of movement, range of shot, damage and so on)
. The state size is 7414, as it is the concatenation of the flattened image-like segmentation map and vectorial inputs.
%The actions available for the agent are, for each hero, moving in any cell within its range of movement, shooting and stabbing. The agent also has the "end of turn" action, it sums up to an action space of size 1258.
Regarding the action space, there are three main group of actions per hero : movement, long and short range attacks, super use. To move, the agent selects a cell on which to send a hero, assuming this cell is within reach of the hero. We use a path-finding algorithm to then move the hero. For the attacks, the agent choose which of the enemies to attack, there is at most 8 enemies. So the total number of possible actions is 3 (number of heroes) x [20x20 (size of the map) + 8 (one shot per enemy) + 8 (one stab per enemy) + 3 (the number of supers)] + 1 (skip) = 1258. Note that not all actions are available for all heroes at all times. For example, the healer cannot use the shield ability or some cells might be out of reach for a hero. When an action is unavailable to the agent, it is simply masked, putting its probability to be selected to 0.

To gather players' data, a play session with 30 participants was organized. We designed a playlist of 10 levels that each player had to play. At the end of each level the player would go on to the next level, no matter the outcome. Playing all 10 levels took each participant roughly one hour of play-time. The game was introduced to them with a tutorial in the form of a PowerPoint document as well as a small video demonstration. 
No further interaction with the players took place to ensure that each player had the same level of information going in. The players only knew that this play session was done in order to help automated test, there was no mention of ML. Most participants are not familiar with RL or with ML in general.
Out of the 30 players, the data of 25 of them was used: the rest either did not play all the way to level 10 or skipped some levels. The first level was removed from the available data, as it mainly served as an introduction level where players where mostly testing the controllers and not engaging fully with the game. Levels 2 through 8 were kept as a training set ($L_{Train}$) and levels 9 and 10 as test levels ($L_{Test}$), which means the training set only contains 7 (levels, $L_{Train}$) * 25 (players, $N_P$) = 175 data points.

\subsection{Training Procedure}

To train the agent, we chose to use the ACER \cite{ACER} algorithm. There are a few reasons why ACER was chosen. First, it is a discrete action algorithm which suits the problem well. It is an on-policy and an off-policy algorithm, allowing for both fast convergence and better use of the data generated.
The off-policy part is coupled with a replay buffer, which is prioritized following \cite{schaul2016prioritized}.
Another major reason for choosing ACER is the possibility to run multiple environments in parallel in an asynchronous fashion, all feeding the same buffer and training the same model. This is quite useful for training agents with an environment that is not perfectly stable and could crash.
The inputs of the agent are both vectorial and convolution-based. The actions of the agent are also both vectorial (shots and stabs) and convolution-based (the movements of each hero). Therefore the neural architecture used was very similar to the one developed in \cite{Quentin_Catane}.
The neural network architecture treats the image-like inputs using 2D convolution layers and the vectorial inputs using dense layer. The features generated are then combined to produce both convolutional (for the movements) and vectorial (for the attacks, supers and skip) outputs for the actions.

Some objectives $z$ are easier than others to reach. For example, shooting as little as the players who shot the least is easier than shooting as much as the players who shot the most. This is the reason why a curriculum-based approach was used to sample the play-modes objectives $z$ at each episode. Moreover, automatic curriculum approaches can improve performances of multi-goal agents \cite{ACL_survey} .
The approach used here is the modeling of absolute learning progress with Gaussian mixture models (ALP-GMM) developed by \citeauthor{ALP_curriculum}.

Realistically, training a RL agent on only 7 levels is usually not enough for the agent to be able to generalize well.
Moreover, in most video games it is usually possible to create more levels, simply by changing the topology, the enemies team composition, or the characters' stats (e.g. health or damage).
%Moreover, in most video games it is usually possible to create randomly generated levels for the agent to train on.
In this work,  the low amount of levels used to train is not due to the low number of levels available to the RL framework, but rather due to the low number of levels the players played on.
%In this work, the low amount of levels used to train is due to the low number of levels the players played on, not necessarily to the low number of levels available to the RL framework.
Finding a way to incorporate additional levels, even without human data, into the training procedure should be a focus in future work. 

Three environments in parallel (each participating in the training of the same model) were used, each running around 12,500 episodes, training the same model. It is equivalent to 24 hours given our computational setup, which is a reasonable constraint to aim for, in real-life use-case of game production.
Our setup is a single computer with a 12 core CPU and a NVIDIA GeForce GTX 1070 GPU.

\section{Evaluation And Results}

\begin{table*}[]

\centerline{
\addtolength{\tabcolsep}{-1.75pt}

\begin{tabular}{|c|c|ll|ll|ll||lll|}
\hline
\multirow{2}{*}{}   & \multirow{2}{*}{\textbf{Metrics}} & \multicolumn{2}{c|}{\textbf{Cluster 1}}                                   & \multicolumn{2}{c|}{\textbf{Cluster 2}}                                   & \multicolumn{2}{c||}{\textbf{Cluster 3}}                                   & \multicolumn{3}{c|}{\textbf{All}}                                                                                 \\ \cline{3-11} 
                                &                                   & \multicolumn{1}{c}{\textbf{Player}} & \multicolumn{1}{c|}{\textbf{CARMI}} & \multicolumn{1}{c}{\textbf{Player}} & \multicolumn{1}{c|}{\textbf{CARMI}} & \multicolumn{1}{c}{\textbf{Player}} & \multicolumn{1}{c||}{\textbf{CARMI}} & \multicolumn{1}{c}{\textbf{Player}} & \multicolumn{1}{c|}{\textbf{CARMI}} & \multicolumn{1}{c|}{\textbf{WinOnly}} \\ \hline
\multirow{8}{*}{\rotatebox[origin=c]{90}{\textbf{Train}}} & \textbf{Stabs}                    & 0.6 (±0.1)                          & 0.6 (±0.1)                          & 1.3 (±0.1)                          & 1.1 (±0.1)                          & 0.9 (±0.2)                          & 0.9 (±0.1)                          & 1.0 (±0.1)                          & \multicolumn{1}{l|}{0.9 (±0.0)}     & 1.1 ( ±0.0)                           \\
                                & \textbf{Shots}                    & 1.7 (±0.1)                          & 1.5 (±0.1)                          & 2.1 (±0.1)                          & 1.9 (±0.0)                          & 1.7 (±0.2)                          & 1.6 (±0.1)                          & 1.9 (±0.1)                          & \multicolumn{1}{l|}{1.7 (±0.0)}     & 0.9 ( ±0.0)                           \\ \cline{2-11} 
                                & \textbf{Empower}            & 0.2 (±0.1)                          & 0.1 (±0.0)                          & 0.8 (±0.1)                          & 0.7 (±0.0)                          & 0.5 (±0.1)                          & 0.4 (±0.1)                          & 0.5 (±0.1)                          & \multicolumn{1}{l|}{0.5 (±0.0)}     & 0.0 ( ±0.0)                           \\
                                & \textbf{Heal}                     & 0.3 (±0.0)                          & 0.2 (±0.0)                          & 0.3 (±0.0)                          & 0.2 (±0.0)                          & 0.1 (±0.0)                          & 0.0 (±0.0)                          & 0.2 (±0.0)                          & \multicolumn{1}{l|}{0.2 (±0.0)}     & 0.2 ( ±0.0)                           \\
                                & \textbf{Shield}                   & 0.1 (±0.0)                          & 0.1 (±0.0)                          & 0.2 (±0.0)                          & 0.1 (±0.0)                          & 0.1 (±0.0)                          & 0.1 (±0.0)                          & 0.1 (±0.0)                          & \multicolumn{1}{l|}{0.1 (±0.0)}     & 0.2 ( ±0.0)                           \\ \cline{2-11} 
                                & \% Win                            & 75 (±11)                            & 52 (±5)                             & 95 (±4)                             & 77 (±3)                             & 87 (±10)                            & 67 (±7)                             & 88 (±4)                             & \multicolumn{1}{l|}{68 (±2)}        & 33 ( ±1)                              \\
                                & \% Lost                           & 0 (±0)                              & 22 (±4)                             & 0 (±0)                              & 18 (±3)                             & 0 (±0)                              & 20 (±6)                             & 0 (±0)                              & \multicolumn{1}{l|}{19 (±2)}        & 8 ( ±1)                               \\
                                & \% Draw                           & 24 (±11)                            & 25 (±4)                             & 4 (±4)                              & 4 (±1)                              & 12 (±10)                            & 12 (±5)                             & 11 (±4)                             & \multicolumn{1}{l|}{12 (±2)}        & 58 ( ±2)                              \\ \hline \hline
\multirow{8}{*}{\rotatebox[origin=c]{90}{\textbf{Test}}}  & \textbf{Stabs}                    & 1.3 (±0.4)                          & 0.9 (±0.2)                          & 2.1 (±0.9)                          & 1.3 (±0.2)                          & 1.4 (±2.2)                          & 1.4 (±0.2)                          & 1.7 (±0.4)                          & \multicolumn{1}{l|}{1.2 (±0.1)}     & 1.6 ( ±0.1)                           \\
                                & \textbf{Shots}                    & 2.1 (±0.2)                          & 2.0 (±0.2)                          & 2.2 (±0.3)                          & 2.2 (±0.1)                          & 1.9 (±0.6)                          & 1.8 (±0.2)                          & 2.1 (±0.2)                          & \multicolumn{1}{l|}{2.1 (±0.1)}     & 1.2 ( ±0.1)                           \\ \cline{2-11} 
                                & \textbf{Empower }            & 0.2 (±0.1)                          & 0.2 (±0.1)                          & 0.8 (±0.3)                          & 0.6 (±0.1)                          & 0.7 (±0.4)                          & 0.3 (±0.1)                          & 0.5 (±0.2)                          & \multicolumn{1}{l|}{0.5 (±0.1)}     & 0.0 ( ±0)                             \\
                                & \textbf{Heal}                     & 0.3 (±0.1)                          & 0.2 (±0.0)                          & 0.3 (±0.1)                          & 0.2 (±0.0)                          & 0.1 (±0.1)                          & 0.1 (±0.0)                          & 0.2 (±0.1)                          & \multicolumn{1}{l|}{0.2 (±0.0)}     & 0.1 ( ±0.0)                           \\
                                & \textbf{Shield}                   & 0.1 (±0.1)                          & 0.1 (±0.0)                          & 0.2 (±0.1)                          & 0.1 (±0.0)                          & 0.1 (±0.1)                          & 0.1 (±0.0)                          & 0.2 (±0.1)                          & \multicolumn{1}{l|}{0.1 (±0.0)}     & 0.5 ( ±0.0)                           \\ \cline{2-11} 
                                & \% Win                            & 100 (±0)                            & 75 (±13)                            & 100 (±0)                            & 80 (±8)                             & 75 (±25)                            & 52 (±20)                            & 95 (±8)                             & \multicolumn{1}{l|}{74 (±7)}        & 33 ( ±5)                              \\
                                & \% Lost                           & 0 (±0)                              & 21 (±13)                            & 0 (±0)                              & 13 (±7)                             & 0 (±0)                              & 43 (±20)                            & 0 (±0)                              & \multicolumn{1}{l|}{20 (±6)}        & 0 ( ±0)                               \\
                                & \% Draw                           & 0 (±0)                              & 2 (±2)                              & 0 (±0)                              & 6 (±5)                              & 25 (±25)                            & 4 (±4)                              & 4 (±8)                              & \multicolumn{1}{l|}{4 (±3)}         & 66 ( ±5)                              \\ \hline
\end{tabular}
}
\caption{Mean and 95\% confidence interval of key-metrics for each cluster for the players and the CARMI agent, on the Train and Test Levels. In bold the metrics used to train the CARMI agent and are reported as the average per turn.}{}
\label{tab:results_all_clusters}
\end{table*}

%The CARMI agent is trained on 2 metrics: the average number of shots per turn and the average number of stabs per turn. Both these metrics lies inside the $[0,3]$ interval, simply because during each turn, each of the three heroes has a maximum of 1 shot and 1 melee attack possible (they can do both). We focus solely on these two metrics because they capture well the aggressiveness of a play-style. Further work should be done in the future to include more metrics, specifically ones measuring the defensiveness of a play-style.

The CARMI agent is trained on 5 metrics: the number of shots, the number of stabs, the number of shots under an empower, the number of heal made and the number of shield used. All these metrics are expressed as a number per turn, so not the number of shots, but the number of shots per turn for example. The first two metrics represent the overall attack strategy of the play-style, while the other three represent the use of the \textit{super capacity} available.

%Additionally we will compare our agent to the CARI approach. Using notations introduced in Section \ref{sec:Human_Like_Behavior}, this CARI agent was trained with $\theta = \{Shot, Stab\}$ and $W = \{[-0.5,2.5], [-0.5,1.5]\}$, so as to match the metrics used to train CARMI.
We will measure the quality of our agent on two aspects : 
%\noindent The claims made about CARMI are twofold:
\begin{enumerate}
    \item %The agent coverage in the play-mode space encompasses the players.
    % The agent is designed to cover at least the same space as the players in the play-modes space
    Does the space of play-modes generated by the agent cover well the distribution of human players ? 
    %\item The agent is configurable with respect to any normalized target metrics
    \item Is the agent capable of generating human-like metrics on new and unseen levels
\end{enumerate}
%In this section, the veracity of these two claims will be measured.

Three models are trained : CARMI, CARI and WinOnly. All three models train on the same levels, have the same state and action space and use the same neural network architecture.

The coefficients of the reward function used by CARI are sampled uniformly in intervals including both positive and negative values, to have the possibility to both encourage and discourage the agent to perform certain actions. This baseline measures what diversity driven agent produces, without taking into account the closeness between its diversity and the diversity in the players play-styles.
The WinOnly model has a sparse binary win/loss reward function. This serve as a baseline to how useful this most commonly used approach would be to give feedback to designers on difficulty assessment, using only a win driven agent, without taking into account the diversity of play-styles.

\subsection{Play-Style Coverage}\label{sec:PS_coverage}
\begin{figure}[h]
%\centerline{\includegraphics[width=0.7\linewidth]{coverage.png}}
%\centerline{\includegraphics[width=\linewidth]{coverage.png}}
\centerline{\includegraphics[scale=0.3]{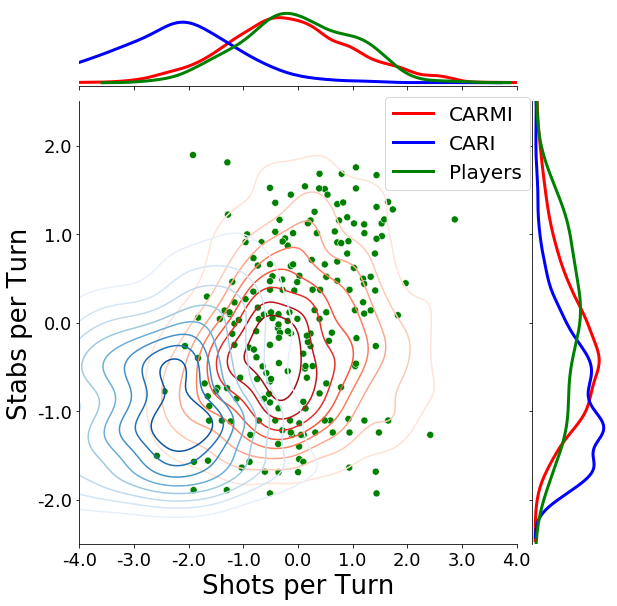}}
%\floatfoot{The metrics here are normalized by the players distribution, each point corresponding to an episode.}
\caption{Coverage of CARI, CARMI and the players. The metrics here are normalized by the players distribution, each point corresponding to an episode.}
\label{fig:coverage_comparison}
\end{figure}

In this section we report the coverage of play-styles for both the CARI and the CARMI agent, and compare them with the players.
We report the results solely on two metrics, due to space constraints: the number of shots per turn and the number of stabs per turn. The CARI agent used here was trained on the same metrics as the CARMI agent.
%Using notations introduced in Section \ref{sec:Human_Like_Behavior}, this CARI agent was trained with $\theta = \{Shot, Stab\}$ and $W = \{[-0.5,2.5], [-0.5,1.5]\}$, so as to match the metrics used to train CARMI.
To measure the coverage possible by the CARI and the CARMI agent we ran 2500 episodes with random $w \sim \mathcal{U}(W)$ for CARI and $z \sim \mathcal{N}(0,1)$ for CARMI.
%To measure the coverage possible by both the CARI and the CARMI agents one must first define the sampling strategy used. In other words: "How to select the $w$ and the $z$ used, respectively by CARI and CARMI ?". %Here in both cases the sampling is made using a uniform distribution.
%For CARI, $w \sim \mathcal{U}(W)$ so as to measure the sub-space "accessible" to CARI in the play-style space.
%For CARMI, $z \sim \mathcal{N}(0,1)$, doing so ensures that we sample most of the values $z$ was trained on.
%95\% of values $z$ was trained on, since $P( -1.96< X \sim \mathcal{N}(0,1) < 1.96) = 0.95$.
%The clustering on the normalized human data was performed and the distribution of each cluster on the normalized play-mode space is reported.
%Moreover, the CARMI agent played 10,000 episodes, each with a randomly uniformly sampled target metrics. The possible coverage space of the agent can therefore be compared with the players one.
These results are reported in Fig \ref{fig:coverage_comparison}.
Additionally, we report the Kullback-Leibler (KL) divergence and the Jensen-Shannon (JS) divergence between each of the models (CARMI, CARI and WinOnly) and the Players, both on the train and on the test levels, on the joint distribution of all metrics in Table \ref{tab:kl_JS}.

\begin{table}[]
\centerline{
\begin{tabular}{|ll|l|l|l|}
\hline
                                             &    & \textbf{WinOnly} & \textbf{CARI} & \textbf{CARMI} \\ \hline
\multicolumn{1}{|l|}{\multirow{2}{*}{\textbf{Train}}} & KL & 10.17   & 9,34 & \textbf{6,69}  \\ 
\multicolumn{1}{|l|}{}                       & JS & 0.76    & 0,73 & \textbf{0,60}  \\ \hline \hline
\multicolumn{1}{|l|}{\multirow{2}{*}{\textbf{Test}}}  & KL & 9,53    & \textbf{9,38} & 9,74  \\ 
\multicolumn{1}{|l|}{}                       & JS & 0,80    & 0,80 & \textbf{0,78}  \\ \hline
\end{tabular}
}
\caption{All metrics joint normalized distribution divergence between players and agents}{}
\label{tab:kl_JS}
\end{table}

The CARMI agent is indeed very much capable to cover the same space as the players. It even generates play-styles not seen in the players population. This too can be an interesting feedback to give to designers as to what is achievable in these levels.

%\begin{figure}[htbp]
%\centerline{\includegraphics[width=\linewidth]{samples/Shot_Dash_Cluster_elipse_normed.png}}
%\floatfoot{\small{The metrics here are normalized by the players distribution, each point corresponding to an episode.}}
%\caption{Distribution of the clusters and the coverage of the CARMI agent (pale rose)}
%\label{fig:coverage_comparison}
%\end{figure}

\subsection{Human-Like Play-Style Emulation}\label{sec:human_ps_sim}
To measure the capacity of the agent to emulate human-like play-style one must first define the play-styles used. To this end, a Gaussian mixture model (GMM) \cite{Gaussian_mixture_model} clustering is trained on the players training normalized summary data:  $\psi_{L_{Train}} := \{\psi_l(\tau_{l,.})\}_{l \in L_{Train}}$, which yields $C$ clusters, each representing a portion of the players. In other words, $\{ \mu_c, \Sigma_c\}_{c \in [1:C]} \gets GMM(C;\psi_{L_{Train}})$. These clusters are what is being used to sample the adequate $z$ in order to emulate players' play-styles..
%and $w$ for CARI.

For CARMI, a number of targets per level is sampled from the distribution of each cluster.
For each level $l \in [1: N_L]$ (including $L_{Test}$) and each cluster $c \in [1:C]$, $z$ is sampled following $\mathcal{N}(\mu_c, \Sigma_c)$ and the episode is run using the trained policy $\pi_z$ on level $l$.
Using this sampling method, the absolute metrics generated by CARMI are compared to what is observed in the players' data, on the train and test levels.

Note that the 7 train levels are used to fit the clustering GMM algorithm. The 2 test levels are used for testing. The human data available on these 2 test levels is not used to re train the clustering or the CARMI agent. So, in effect, neither the agents, nor the clustering, use the test levels for anything else than evaluation.
%For CARI, we use the same process with the addition of a linear function ($f)$ used to bridge the gap between the desired $z$ and the according $w$ used to condition the policy. We fit this function using the dataset generated in Section \ref{sec:PS_coverage}. $f$ takes as input a normalized summary data (i.e. $\psi_l(\tau_{\pi_{CARI}})$) and outputs the corresponding reward coefficients which generated $\tau_{\pi_{CARI}}$ (i.e. $w$). For each level $l \in [1: N_L]$ (including $L_{Test}$) and each cluster $c \in [1:C]$, a $z$ is sampled following $\mathcal{N}(\mu_c, \Sigma_c)$ and the episode is run using the trained CARI policy $\pi_{f(z)}$ on level $l$.

We do not compare results with CARI here for 2 reasons. First, there is no straightforward way to transform the samples $z$ from the clusters into $w$ to feed the CARI agent. Second, even if there were, it is obvious from the big gap in Fig. \ref{fig:coverage_comparison} between the CARI and the players, that the CARI agent would never be able to emulate the players accurately. We do however, compare those results with a pure RL model trained with the sole objective of winning, called "WinOnly". This model is here to compare with what is possible using "\textit{classical}" RL. Indeed, training a model solely to win and using it to inform designers on balancing issues is not unheard of in the industry.
Note that neither the training of CARMI nor the training of the clustering have seen the test levels nor the players' data on those levels, these are truly previously unseen levels. These results are available in Table \ref{tab:results_all_clusters}.

%The first thing to notice is the capacity of the agent to match the desired metrics on the train levels, it is a good indicator that the model did converge. Also it is worth noting that the agent is well capable to generalize those play-styles to unknown levels, in the test set. 
%While the metrics do not match perfectly between the players and the agent, they are however ordered in the same fashion. Indeed, for both the players and the CARMI agent: cluster 1 stabs the least, cluster 2 has a frequency of use for both knife and gun quite average and cluster 3 uses the knife and the gun a lot more than the other players. 
%Another very interesting result is the win rate. Indeed 

The first thing to notice is the capacity of the agent to match the desired metrics on the train levels, it is a good indicator that the model did converge. Also it is worth noting that the agent is overall capable to generalize those play-styles to unknown levels, in the test set.
Looking at what the CARMI agent has produced, the feedback that we would have given the designers would have been that the new levels will have the players shoot and dash more, but that the use of the \textit{super capacities} would remain somewhat stable. These feedback would have been correct since they match what is observed in the player's data for the test levels. 
%For each of the play-modes encoding the play-styles (in bold), whether on train or test levels, CARMI has its clusters ordered the same way as the players. This shows that the agent has learned to produce metrics matching the way we encoded the play-style (i.e. $\psi_l(\tau)$ instead of $\psi(\tau)$).
%While most absolute metrics are quite close between CARMI and the players on test levels (e.g. the number of shots per turn), it is worth noting that some metrics however, are not as similar (e.g. the number of stabs per turn). This might be solved using more data, or maybe changing the sampling strategy by removing the assumption that $\psi(\tau_{l,.}) \sim \mathcal{N}(\mu_{l}, \sigma_{l})$ by using other probability distributions.
Had we used the WinOnly model to give feedback to the designers we would have been right about the shots and dash, we would however have missed completely the feedback on the \textit{super capacities}. This goes to show that encouraging models to fit the players play-modes allows to create mode meaningful feedback from automated tests.
%With the exception of the number of stabs per turn which is quite low compared to the players. However, the number of shots per turn increases in the test levels both for the players and the CARMI agent.
%In the absence of players data, we could have safely used our agent to inform the designers that these new levels will have the players shoot more than average.
%A similar conclusion can be drawn on the \textit{super capacities}. Indeed, looking at the heal and shield we would have given the feedback to the designers that the new levels would not drastically change the frequency of use of these 2 \textit{super capacities}: they remain the same between train and test, both on players and agent. However, the number of shots made under empower does increase a little bit on the new levels.
%Overall it is noticeable that the differences between the players and the agent are bigger at the clusters level rather than at the population level.

The other very interesting result are the win rates. Overall the players perform better on the new level: from 88\% to 95\% of game won. The same is observed with the agent: from 68\% to 74\%. But even more interesting are the win rates of the clusters between the train and the test levels. The players in clusters 1 and 2 both perform better in the test levels, and this increase in performance is mirrored by the agent emulating these two clusters. However the players in cluster 3 perform worse: from 87\% to 75\%. The agent, when emulating this third cluster, also performs worse. 
Indeed, regarding difficulty assessment, given what the CARMI agent has produced, we would have concluded to the designers that the new levels are overall easier, expect for the third cluster.
%Moreover, again looking at what CARMI has produced, we can even hypotheses, by comparing the clusters emulated by CARMI on test levels, that this decrease in performance for the third cluster is due to a low use of the rifle and of the heal.
Additionally, as for the play-modes, the win rates of the WinOnly model is not reliable and doesn't allow to draw correct conclusions.

\section{Conclusion And Discussion}

We have developed a new agent capable, with one single training phase, to generate a continuum of play-styles which includes the players' ones, using limited human summarized data. We have also developed and demonstrated the effectiveness of a straightforward sampling strategy able to generate human-like play-styles on new levels reliably. This approach can provide very meaningful feedback to level designers.
This approach makes no assumption as to the type of environment, or RL algorithm used, making it easily usable in many different contexts.

Improvements should be made to incorporate more play-modes in order to capture more of the players play-styles. We argue that the more numerous the play-modes, the smaller the differences in win rates.
%Automating the selection of these play-modes would go a long way towards the adoption of this approach.
Another limitation of this approach is the number of level used for training being limited by the available human data. Including randomly generated levels (or at least some variations of the training levels), lacking human data, into the training procedure would also allow the agent to generalize better between the train and the test levels. 
%This might be done using a mix of CARI and CARMI.
Furthermore, we notice that the differences between the agent and the players are bigger at the level of the clusters rather than at the level of the whole population. Including the players' clusters distribution in the learning phase would probably increase the accuracy of the agent when emulating each of these clusters. Simply put, using the clustering during both learning and inference would allow a better approximation of the clusters by the agent.

Moreover, this method relies on metrics computed over a whole episode (i.e. a whole game). So, for games with very long episodes (e.g. RTS games) there are many ways to reach the desired metrics. This might produce unexpected results. For example, in a RTS game with a metric measuring the amount of resources gathered, the CARMI agent could gather the desired amount of resources in an unexpected way. This could indeed produce misleading feedback for the designers. One way to counter balance this is to increase the number of target metrics used and their variety to capture more gameplay aspects. The more numerous the number of metrics, the more constrained the agent should be. Studying the effect of a varying number of metrics should be done in the future.

% Use \bibliography{yourbibfile} instead or the References section will not appear in your paper
%\small{\bibliography{aaai22}}

%\fontsize{9.8pt}{10.8pt} \selectfont

%\fontsize{10.8pt} \selectfont
\normalsize
\appendix
\section{Additional Results}
%In this section we  display additional results.

\subsection{Play-Modes Distributions}
\begin{figure}[h]
\centerline{
    \includegraphics[width=.24\textwidth]{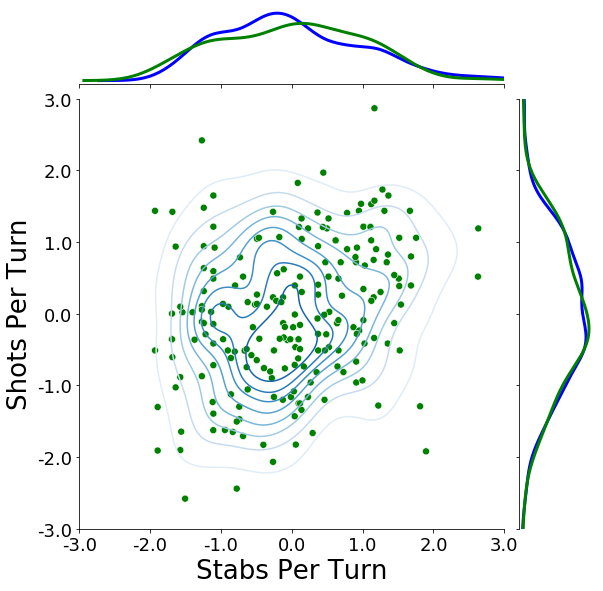}\hfill
    \includegraphics[width=.24\textwidth]{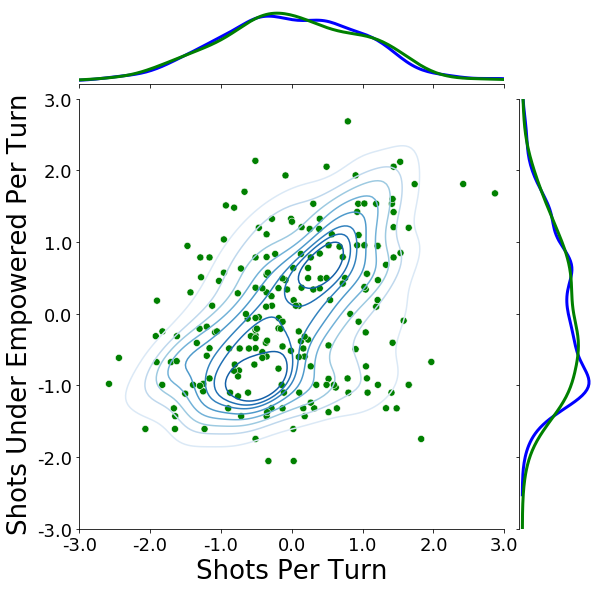}}
    % \\[\smallskipamount]
\centerline{
    \includegraphics[width=.24\textwidth]{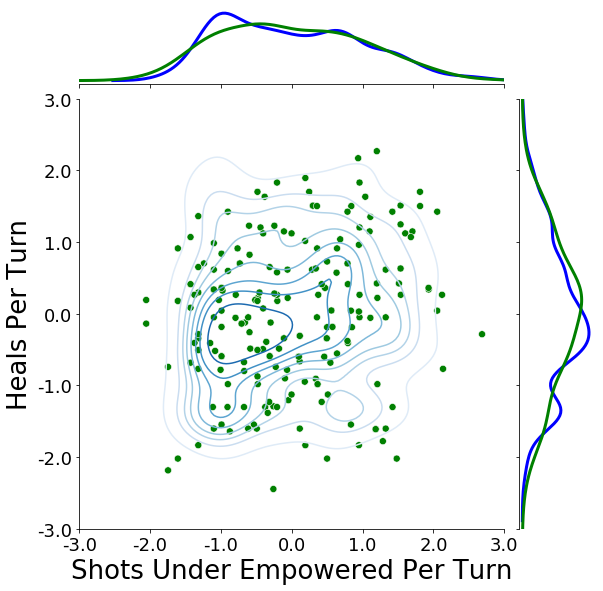}\hfill
    \includegraphics[width=.24\textwidth]{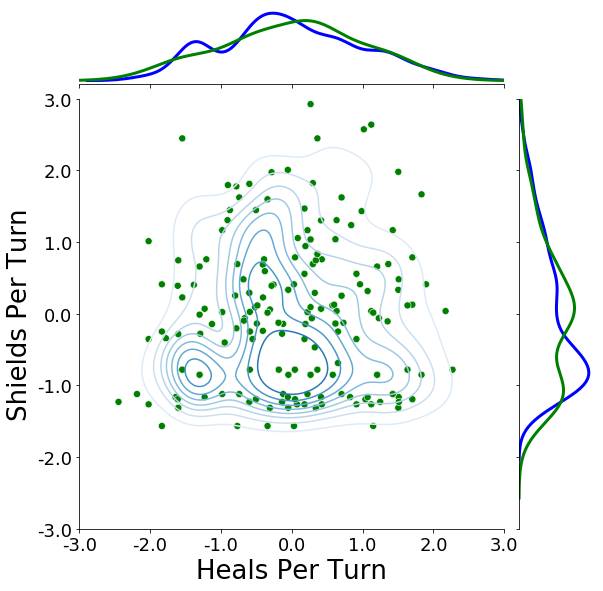}}

    \caption{Distribution over the play-modes of players (in green) and CARMI (in blue) normalized summary data, following the sampling strategy described in section \ref{sec:human_ps_sim}}
    \label{fig:coverage_all_metrics}
\end{figure}

In Table \ref{tab:results_all_clusters}, we reported only the mean and 95\% confidence interval of each play-mode for the players' clusters and their CARMI counterpart. Here we report the full distributions. These results can be seen in Fig. \ref{fig:coverage_all_metrics}. The data that produced Table \ref{tab:results_all_clusters} is the same displayed in Fig. \ref{fig:coverage_all_metrics}. 

These graphs showcase that our model is capable to cover the space of players play-style and that the very simple sampling strategy is adequate to have the agent emulate the players' play-style.

\newpage 
\subsection{Learning Curves}
\begin{figure}[h]
\centerline{
    \includegraphics[width=.4\textwidth]{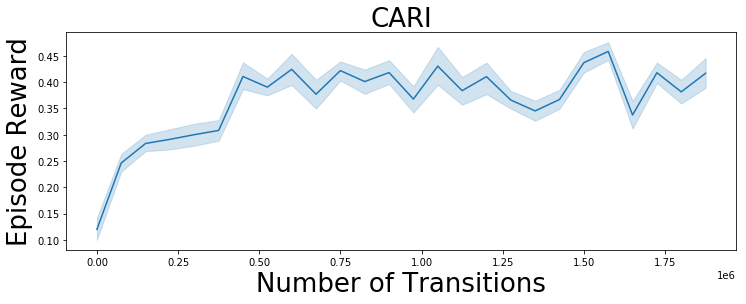}
    }
\centerline{
    \includegraphics[width=.4\textwidth]{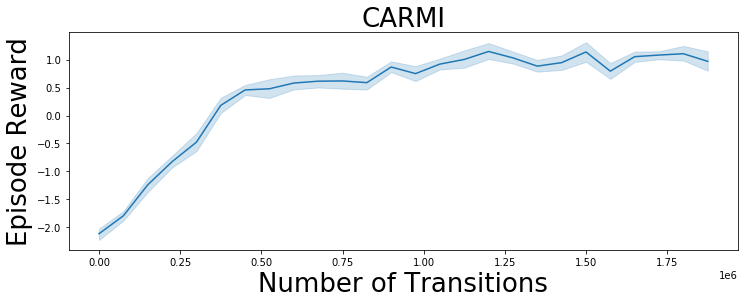}
    }
\centerline{
    \includegraphics[width=.4\textwidth]{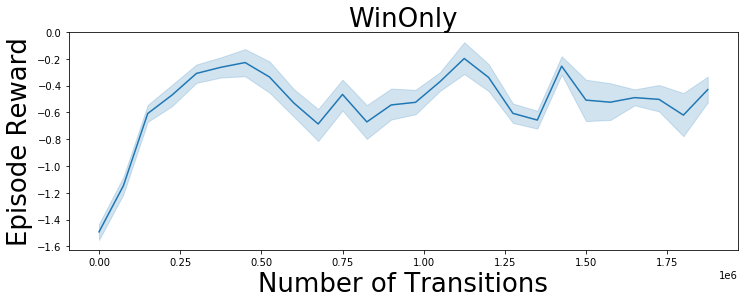}
    }
    \caption{Evolution of the CARI, CARMI and WinOnly agent episodic reward.}
    \label{fig:reward_training}
\end{figure}

We display the evolution of the episodic reward throughout the training for both the CARI and the CARMI agent, in the Fig. \ref{fig:reward_training}.

%It demonstrates that the models did train enough.

\bibliography{main}

\end{document}